%% file: template.tex
\documentclass{vgtc}                          

\graphicspath{{figures/}{pictures/}{images/}{./}} 

\usepackage{times}                     

\usepackage{tabu}                      
\usepackage{booktabs}                  
\usepackage{lipsum}                    
\usepackage{mwe}                       
\usepackage{enumitem}

\usepackage{mathptmx}                  

\usepackage[cjk]{kotex}
\usepackage{mathtools}

\DeclarePairedDelimiter\floor{\lfloor}{\rfloor}

\onlineid{1072}

\vgtccategory{Research}

\vgtcinsertpkg

\title{Assessing Graphical Perception of Image Embedding Models \\ using Channel Effectiveness}

\author{
    Soohyun Lee \thanks{e-mail: shlee@hcil.snu.ac.kr}\\ %
        \scriptsize Seoul National University %
    \and Minsuk Chang\thanks{e-mail: jangsus1@snu.ac.kr}\\ %
        \scriptsize Seoul National University 
    \and Seokhyeon Park\thanks{e-mail: shpark@hcil.snu.ac.kr}\\ %
        \scriptsize Seoul National University 
    \and Jinwook Seo\thanks{e-mail: jseo@snu.ac.kr}\\ %
        \scriptsize Seoul National University 
}

\abstract{

\input{sections/00-Abstract}
} 

\keywords{Graphical perception, channel effectiveness, image embeddings, clip}

\begin{document}


\firstsection{Introduction}
\maketitle

\input{sections/01-Introduction}
\input{sections/02-Related_Work}

\input{sections/03-Linearity_Analysis}

\input{sections/04-Discriminability}

\input{sections/05-Applications}

\input{sections/06-Conclusion}

\acknowledgments{
    This work was supported by the National Research Foundation of Korea (NRF) grant funded by the Korean government (MSIT) (No. 2023R1A2C200520911)
}

\bibliographystyle{abbrv-doi-hyperref}

\bibliography{template}
\end{document}

%% file: sections/00-Abstract.tex
Recent advancements in vision models have greatly improved their ability to handle complex chart understanding tasks, like chart captioning and question answering. However, it remains challenging to assess how these models process charts. Existing benchmarks only roughly evaluate model performance without evaluating the underlying mechanisms, such as how models extract image embeddings. This limits our understanding of the model’s ability to perceive fundamental graphical components.
To address this, we introduce a novel evaluation framework to assess the graphical perception of image embedding models. For chart comprehension, we examine two main aspects of channel effectiveness: accuracy and discriminability of various visual channels. Channel accuracy is assessed through the linearity of embeddings, measuring how well the perceived magnitude aligns with the size of the stimulus. Discriminability is evaluated based on the distances between embeddings, indicating their distinctness.
Our experiments with the CLIP model show that it perceives channel accuracy differently from humans and shows unique discriminability in channels like \textit{length, tilt, and curvature.} We aim to develop this work into a broader benchmark for reliable visual encoders, enhancing models for precise chart comprehension and human-like perception in future applications.

%% file: sections/01-Introduction.tex

Nowadays, emerging vision models are strongly influencing the domain of visualization, especially in handling charts. Image encoders are employed to automatically classify chart images~\cite{chartClassification,chartClassification2}, explain charts~\cite{chartCaption,chartCaptionBenchmark}, or answer chart-based questions~\cite{chartllama, chartqa}.
This is due to the recent advancements in vision models' ability to process visual data and perform diverse tasks (e.g., saliency prediction~\cite{saliency}, image captioning~\cite{clip, caption}, or visual question answering~\cite{vqa}), often surpassing human-level performance~\cite{objectDetectionSurpass, classificationSurpass}.

However, most existing benchmarks for chart understanding models focus on high-level tasks, such as task performance in question answering~\cite{chartqa, chartbench} or image captioning~\cite{chartCaptionBenchmark} scenarios. 
These benchmarks can evaluate the model's overall performance and utility but are too coarse to address how they perceive and interpret the fundamental graphical elements in charts at a perceptual level. 

To address this gap, we introduce a novel evaluation framework for image embedding models based on the concept of `channel effectiveness,' which considers two main aspects: accuracy (\autoref{sec:linearity}, \autoref{sec:discriminability}) and discriminability(\autoref{sec:discriminability}).
Our framework can measure how precisely vision models can interpret and discriminate the magnitude channel typically used in charts~\cite{automating, munzner}: \textit{length}, \textit{tilt}, \textit{area}, \textit{color luminance}, \textit{color saturation}, and \textit{curvature}.

First, we suggest using the linearity of image embeddings as a proxy for each channel's accuracy while their magnitude increases. 
According to Steven’s power law, channel accuracy improves as the perceived magnitude increases linearly with the given stimuli.
Then, we broadened our investigation of whether the order of measured linearity could be generalized across all combinations of controlled variables. 
Secondly, we suggest analyzing the distances between consecutive embeddings to evaluate each channel's discriminability.
We calculated the peaks from the smoothed distance graph to ascertain the number of distinguishable groups and how sensitive the model reacts to the magnitude of the channel.

We also present that our evaluation framework can measure the low-level performance of the model depending on the model's goal:
\begin{itemize}[left=4pt, itemsep=0pt]
    \item[\textbf{1.}] \textbf{Tasks requiring precise quantitative analysis} (e.g., determining exact values from a bar chart). In such cases, models should give precise answers from the graphical elements. Therefore, they must achieve higher channel accuracy and maintain low discriminability when the magnitude increases.
    \item[\textbf{2.}] \textbf{Tasks where models need to process charts as humans do} (e.g., interpreting trends from a line graph~\cite{diversingPerspectives} or mimicking user studies~\cite{gptUserStudy}). In this scenario, the accuracy should align with the known perceptual effectiveness of human vision~\cite{channelMturk, GraphicalPerception} (should follow human perceptual ranking), and discriminability should mirror human ability~\cite{munzner, discriminabilitytest}.
\end{itemize}

We applied our framework on CLIP~\cite{clip}, one of the state-of-the-art image embedding models pre-trained on a large-scale dataset of natural images.
The result reveals that CLIP's order of channel accuracy differs from human perception and each accuracy is much lower than being ideally linear.
Furthermore, CLIP exhibits unique discriminability patterns that seem to follow human perception on certain channels, such as length, tilt, and curvature.
We also found that CLIP's perception conforms to Weber's law~\cite{weberslaw}, indicating that perceived changes in stimuli are proportionate to the magnitude of the initial stimuli.
Comprehensively, we observed a tradeoff between accuracy and discriminability, where accuracy can be lowered when there exists a certain amount of discriminability.

We present our framework for channel effectiveness as a foundational effort in establishing low-level benchmarks for chart comprehension.
Furthermore, we suggest the visualization community explore additional low-level benchmarks, such as pre-attentive processing~\cite{preattentive} or just-noticeable difference (JND)~\cite{weberslaw}.
Our future initiatives include collecting crowdsourced data to validate our findings from discriminability and comparing them with results from various image embedding models to confirm the robustness and applicability of our benchmark.

%% file: sections/02-Related_Work.tex
\section{Related Work}

\subsection{Graphical Perception}
Cleveland and McGill~\cite{GraphicalPerception} tried to understand encodings in visualization through graphical perception.
They measured humans' graphical perception by defining 10 elementary perceptual tasks (e.g., position, length, angle, area, and volume), collectively known as channels, that people use to extract quantitative information from graphs.
They also performed pairwise experiments, such as comparing bar charts and pie charts to compare the difficulty between position and angle, and then ranked these channel effectiveness based on the accuracy of human perception.

Various research extended this in terms of participant scale~\cite{channelMturk}, data type~\cite{timeseries, timeseries2}, or task complexity~\cite{complexPerception1, complexPerception2}, leading to a broader and more solid understanding of graphical perception.
Furthermore, the emergence of CNN models questioned researchers on whether the same findings also apply to the trained models~\cite{cnnPerception}.

Traditional methods primarily evaluate graphical perception by assessing channel effectiveness in terms of accuracy with human subjects or through models trained with predetermined answers.
However, these approaches cannot be easily applied to computer vision models. Also,  they typically miss examining other critical aspects of channel effectiveness~\cite{munzner} (e.g., discriminability~\cite{clams}, separability, popout, and grouping), which are crucial for understanding how visual information is perceived and processed.

Therefore, our paper introduces a method for understanding how unsupervised image training models perceive channels by analyzing the raw outputs of image embedding models. Beyond accuracy, our investigation includes discriminability, which refers to the ability to differentiate between similar visual elements, illustrating a broader perspective on the components of channel effectiveness.

\subsection{Image Embeddings}
Computer images are structured pixels of light and color intensities, which are not directly related to their intrinsic meaning.
Therefore, various image encoding models were suggested to transform images into embeddings, a fixed-length vector representing their semantics~\cite{flamingo, chartqa, clip, imageTextEmbedding}.
These image embedding models are often used in other models' backbone to classify, describe, interpret, or perform multimodal tasks~\cite{llava, instructblip}.
Also, various benchmarks were suggested to evaluate such models through QA tasks ~\cite{chartbench, chartCaptionBenchmark}.
However, we found that the image embedding itself has neither been investigated nor evaluated. Therefore, we present a methodology to measure graphical perception within these image embeddings and experiment with CLIP~\cite{clip}, a general state-of-the-art image embedding model used in various domains (medicine~\cite{medicalCLIP}, fashion~\cite{fashionCLIP}, or even user interfaces~\cite{uiCLIP}).



%% file: sections/03-Linearity_Analysis.tex


\begin{figure}[t]
    \centering
    \includegraphics[width=\linewidth]{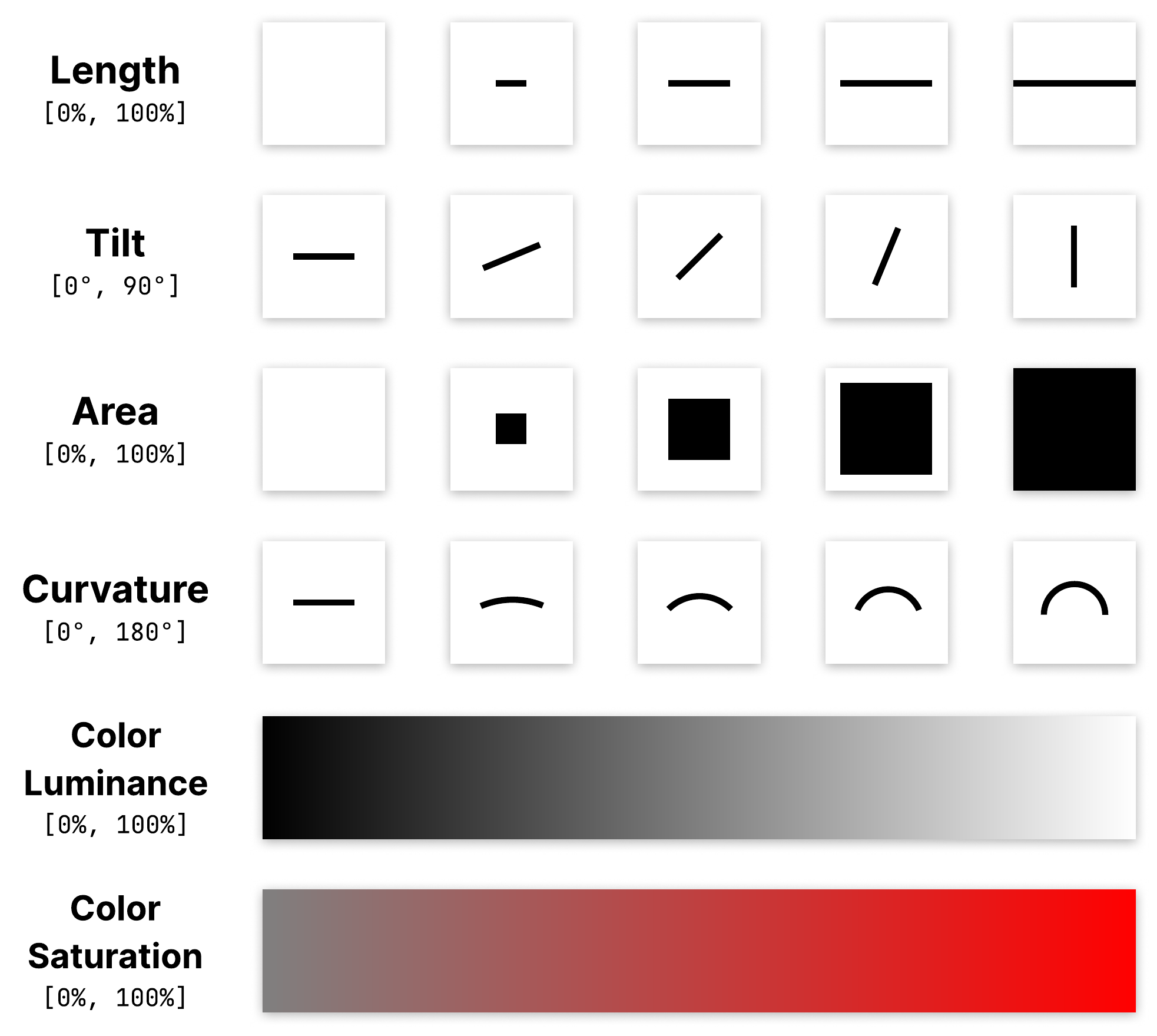}
    \caption{
        Examples of our variations for each channel. \textbf{Length} and \textbf{Area} of 100\% means the line or square fills the screen. \textbf{Tilt} is adjusted from 0$^\circ$ to 90$^\circ$. \textbf{Curvature} starts from a straight line to a semicircular arc. Color Hue is fixed to 0 (red) when the \textbf{Luminance} and \textbf{Saturation} increase from 0\% to 100\%. 
    }
    \label{fig:channel_values}
    \vspace{-1em}
\end{figure}

\section{Linearity as Channel Accuracy} \label{sec:linearity}


In pursuit of understanding how image embedding models like CLIP capture variations on different visual channels, we designed experiments to assess their sensitivity to changes in six different channels: \textit{Length, Tilt, Area, Color Luminance, Color Saturation,} and \textit{Curvature}.
Previous studies have shown that the accuracy of human perception of these visual channels differs by order: length is most accurately perceived, followed by tilt, area, color luminance, color saturation, and curvature~\cite{GraphicalPerception,channelMturk}.
We examined whether an image embedding model can produce embeddings with a strict order of accuracy.

\subsection{Linearity with Fixed Control Variables}
\label{sec:linearity_single}
\subsubsection{Experiment Design}
In our experiment, we generated an image dataset of simple shapes, following the prior work~\cite{channelMturk}.
Our goal was to eliminate any unintended bias from the background, ensuring a focus purely on the graphical perception of the elements.
We first created images of a line segment with each channel applied with a certain magnitude on a white background.
For each channel, we encoded the range of values in \autoref{fig:channel_values} over 1000 steps.
While testing on one channel, other channels are fixed to the controlled (default) value in \autoref{fig:channel_values} (Length: 50\%, Tilt: 0$^\circ$, Area: None, Curvature: 0$^\circ$, Luminance: 50\%, Saturation: 100\%).
For example, in the luminance channel, the line segments with a Length of 50\% and no 
Tilt or Curvature were rendered with varying degrees of brightness.

For each image, we extracted embeddings using three CLIP models~\cite{clip} with different visual architecture; one ResNet~\cite{resnet} model (RN50x64) and two Vision Transformer~\cite{vit} models (ViT-B/32 and ViT-L/14@336px). 
We then analyzed linearity using principal component analysis (PCA)~\cite{pca}, observing how well the first principal component could represent the distribution of embeddings.
This allows us to measure the linearity of each channel's embedding space quantitatively.
We also claim that linearity can be a good proxy for measuring channel accuracy, where the single-scale change in its stimuli directly appears in the embedding space.
Also aligned with that, an accurate channel to humans means that human perception is proportional to stimuli~\cite{powerlaw}.

\subsubsection{Result}

\begin{figure}[t]
    \centering
    \includegraphics[width=\linewidth]{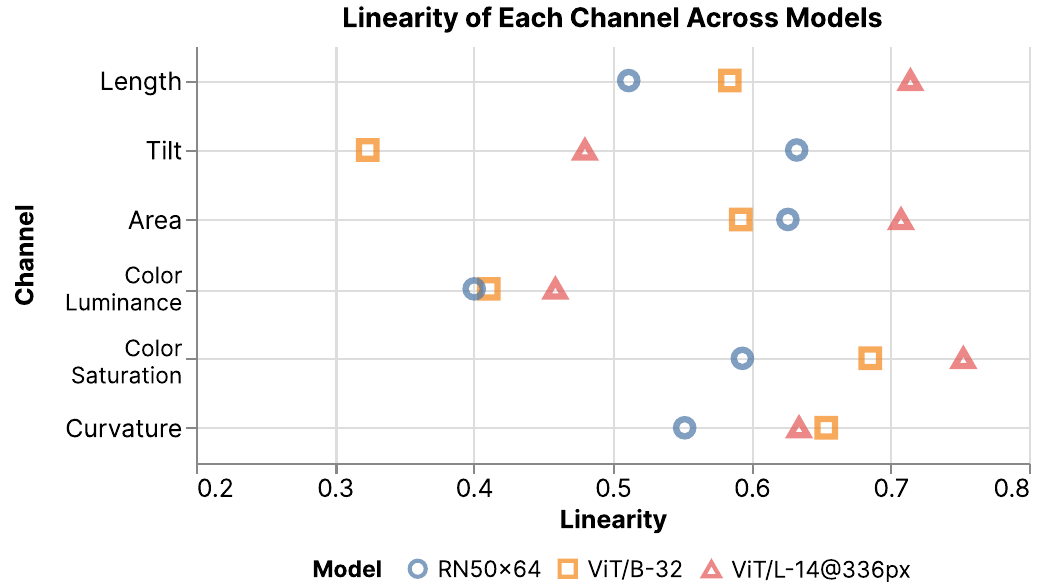}
    \caption{
        Linearity of various visual channels across different CLIP models.
        The Y-axis is the channel arranged in the order that humans perceive more accurately.
        Each channel's linearity varies between models, which does not closely align with human perceptual accuracy.
        Also, the ViT-L/14@336px model usually shows better accuracy compared to other models.    
    }
    \label{fig:linearity_single}
    \vspace{-1em}
\end{figure}

The linearity of each channel for each model is plotted in \autoref{fig:linearity_single}.
On the y-axis of the figure, the channels are arranged in the order that humans perceive more accurately, starting from the top.
We can find that the order does not align with human perception for all models. 
Notably, model ViT-B/32 shows significantly lower linearity in channel tilt than other models.
Additionally, for all models, \textit{color luminance} shows the lowest linearity compared to other channels.
This suggests that important data encoded as \textit{color luminance} can potentially cause the CLIP model to recognize or misinterpret its precise value barely.

\subsection{Linearity with Every Controlled Variables}
We also investigated whether the findings can be generalized into circumstances where other channels are not fixed. This experiment reflects a more complex and realistic scenario similar to natural visual environments where data tends to be encoded in multiple channels simultaneously.

\subsubsection{Experiment Design}
We explored the linearity of channels by examining all possible combinations of channel variations, finding out whether a general order of channel effectiveness exists.
In~\autoref{sec:linearity_single}, we experimented with varying channel magnitude with 1000 steps where other channels are fixed as controlled values in~\autoref{fig:channel_values}.
However, in this extended experiment, we reduced the number of steps to 20 while testing all combinations of other channels, resulting in $20^4$ image variations per channel.
We then measured every channel combination's linearity and observed whether a general order of channel accuracy exists.

\subsubsection{Result}
\begin{figure}[t]
    \centering
    \includegraphics[width=\linewidth]{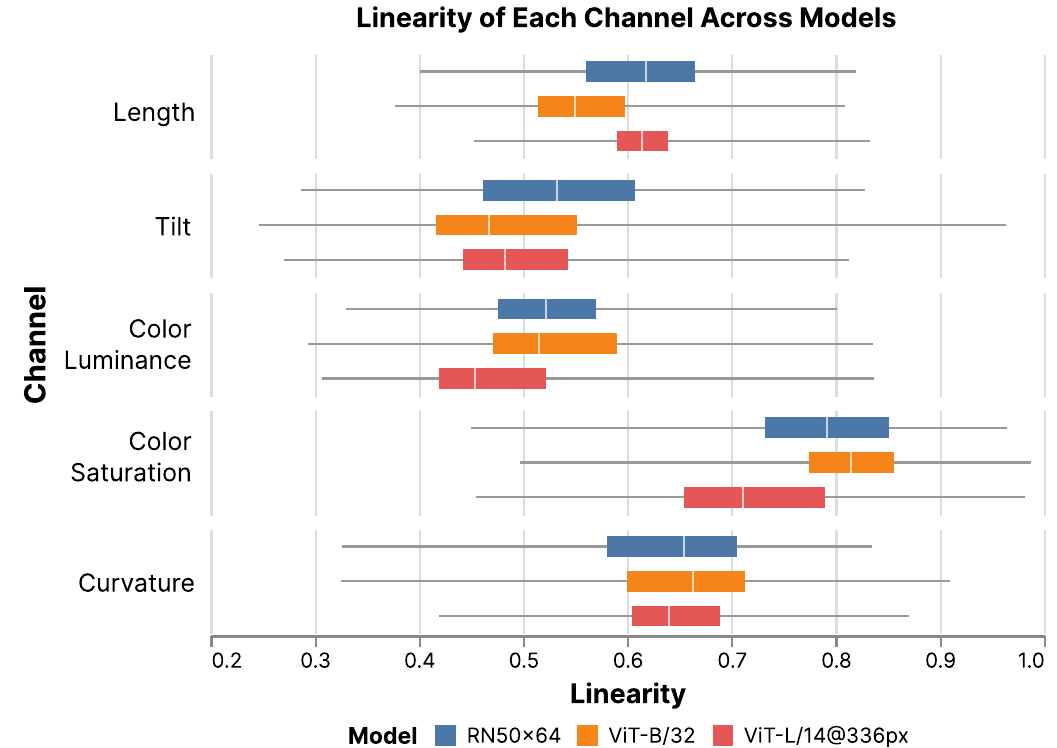}
    \caption{
        The box plot illustrates the linearity scores for each channel under every combination of controlled variables, showcasing general patterns and deviations.
        Since area cannot be applied together with length or curvature, we have generated combinations without area.
        Based on the plot, all models have similar overall rankings for the channels they perceive (\textit{Color saturation $>$ Curvature $>$ Length $>$ Tilt $=$ Color luminance}).
        The whisker of this boxplot represents the min and max of the full data.
    }
    \label{fig:linearity_all}
    \vspace{-1em}
\end{figure}

The calculated linearity scores are plotted in~\autoref{fig:linearity_all}.
We can easily observe a general order of linearity among the channels (\textit{Color~Saturation} $>$ \textit{Curvature} $>$ \textit{Length} $>$ \textit{Color~Luminance} $=$ \textit{Tilt}), which is also similar to the result from~\autoref{sec:linearity_single}.
Furthermore, we conclude that the CLIP model shows a significant difference in channel accuracy compared to human perception. This disproves the assumption that vision models would process the data similarly to humans.

\subsection{Discussion}
The previous two experiments reveal that CLIP's channels' effectiveness ranking differs from that of humans, and some of the channels show low linearity. Different channel rankings indicate that the model perceives images differently from humans, leading to a risk of inaccurate results when using CLIP embeddings to mimic human perceptions. Also, the low linearity channels can cause inaccurate answers to be produced when performing quantitative QA based on graphical elements through CLIP embedding. We suggest these results should be considered when using the image embedding model.

%% file: sections/04-Discriminability.tex
\section{Distance as Channel Discriminability}
\label{sec:discriminability}

\begin{figure*}[t]
    \centering
    \includegraphics[width=1\textwidth]{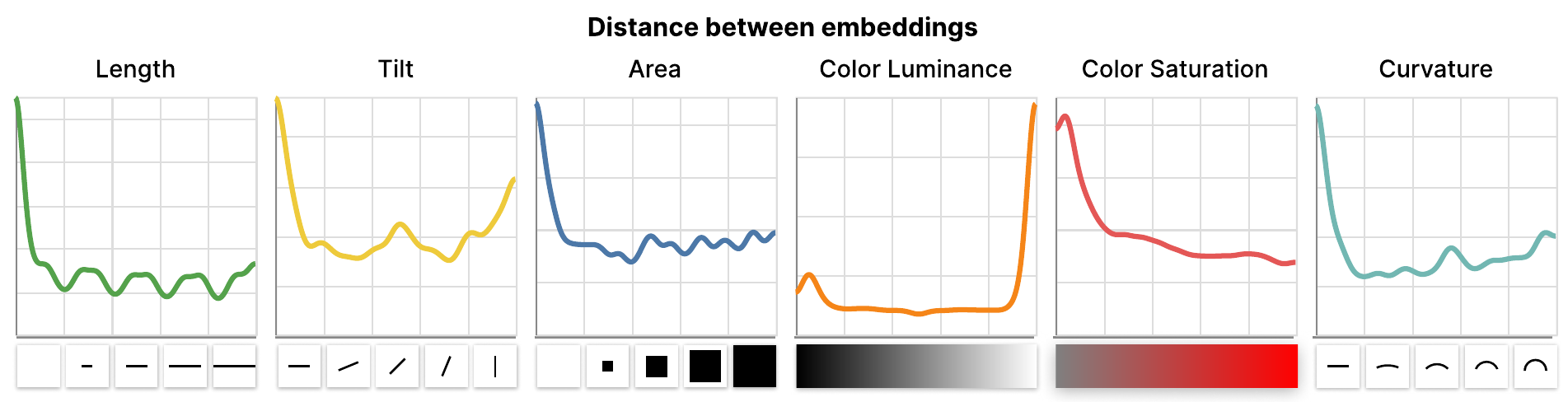}
    \caption{
        The smoothed plot of the Euclidean distances between image embeddings for incremental changes in each visual channel.
        Sample images below the chart are illustrations of stimuli variations in each channel.
        Peaks represent thresholds where the model perceives significant differences between images, indicating the discriminability of each channel.
        This visualization aids in identifying how many perceptual groups the model can distinguish in each channel.
    }
    \label{fig:smooth_distance}
    \vspace{-1em}
\end{figure*}

In this section, we explore the concept of discriminability (one measure of the channel's effectiveness) within the context of image embedding models.
Discriminability refers to the capacity to perceive distinct steps or changes within visual elements of an image.
Conversely, the existence of such discriminability suggests that accuracy may be low.
For instance, it has been observed that distinguishing more than six hues or more than six symbol shapes within a visual array can be challenging, suggesting a perceptual threshold for discrimiability~\cite{automating}.

This threshold indicates the minimum difference required between two objects to be considered distinct.
To assess this, we measured the Euclidean distance between image embeddings, as these distances can be interpreted as the model’s ability to differentiate between two images.

We measured the Euclidean distance between embeddings, calculated with the best-performing model (ViT-L/14@336px), of each consecutive pair in a series of 1000 images, each created by incrementally changing the value of one channel as described in \autoref{sec:linearity_single}.
The results of these measurements are smoothed with a Gaussian filter~\cite{gaussianfilter} ($\sigma = \floor*{\sqrt{1000}} = 32$) and plotted in \autoref{fig:smooth_distance}.
The smoothed plot helps us observe the boundary between the groups the model perceives as different, thereby showing how much change in one channel is necessary before the model perceives the two images as different.
We expect the number of distinct peaks can be a proxy for extracting separable groups throughout each channel. 

\subsection{Result \& Discussion}

The result from this experiment roughly explains how the model discriminates changes across different channels.

\begin{itemize}[label={}, left=0pt, itemsep=0pt]
    \item \textbf{Length:} First, when analyzing the discriminability of the length channel, we can see that when the length is short, the distance between adjacent embeddings becomes relatively high. Supported by Weber's law~\cite{weberslaw}, the model captures subtle changes well when the length is short, similar to humans. Interestingly, the distance graph for Length shows three or four distinct peaks with valleys between, indicating that the model recognizes length in four separate stages.
    
    \item \textbf{Tilt:} The distances are noticeably higher for tilt at angles of 0$^\circ$, 45$^\circ$, and 90$^\circ$. This pattern shows that the model primarily differentiates images based on whether the tilt is less than, equal to, or greater than 45$^\circ$. In other words, we can assume that the model uses 45$^\circ$ as an important threshold for processing images.
    
    \item \textbf{Area:} Upon analyzing the area, several small peaks were identified throughout the range. Unlike length, we cannot easily distinguish or split into distinct groups.
    
    \item \textbf{Color Luminance:} We can observe a small peak near 0\% and an extreme peak at 100\%.
    This indicates that the model is particularly sensitive to changes in luminance at very low and very high values.
    It suggests that when it is very dark or very bright, it responds greatly to small changes in luminance, and in other areas, images are viewed relatively similarly.
    
    \item \textbf{Color Saturation:} Distance for color saturation was high near the value of 0\%. This peak indicates that the model is highly responsive to initial saturation increases in the low saturation state but becomes less sensitive once the change has been made.
    
    \item \textbf{Curvature:} Curvature has very high distance values when approaching 0$^\circ$, showing high sensitivity when there are almost no curves. Also, a peak around 90$^\circ$ appears to be a critical point similar to tilt, which becomes a perceptional boundary.
    
\end{itemize}

%% file: sections/05-Applications.tex
\section{General Discussion \& Future Application}

We proposed an evaluation framework (\autoref{sec:linearity}) that extensively investigated the channel effectiveness of the image embedding model and then applied it to the CLIP model. However, we summarize our limitations and suggest our corresponding future work.

\subsection{Reliable Visual Encoders for Charts}
\label{sec:suggestAcc}
The analysis from~\autoref{sec:linearity} shows that generally trained vision models may interpret the encoded data unreliably, where linearity for channels except \textit{color saturation} stays around 0.6. 
This can be problematic in chart question-answering models where accurate interpretation of visual cues is essential. 
Accurate interpretation requires an accurate perception of channels. Thus, accuracy for all channels should be linear (linearity score close to 1).
Conversely, checking whether the model's discriminability matches human perception can be crucial in chart captioning models. 
Chart captioning requires a holistic understanding of existing visual elements, where matching its discriminability with humans is essential.
Therefore, in this case, the order and intensity should be similar to that of humans besides having higher channel accuracy.
Therefore, we suggest that chart question-answering models should be trained to have higher channel accuracy, while chart captioning models should have similar accuracy as humans.

\subsection{Ambiguity in Peak Analysis for discriminability}

In our analysis of discriminability in \autoref{sec:discriminability}, we conducted peak analysis to investigate the model's effectiveness in differentiating channels.
As discussed in the previous section (\autoref{sec:suggestAcc}), two-directional goals emerge: reporting high accuracy and matching human discriminability.
To interpret charts accurately, the graph shapes for each channel in \autoref{fig:linearity_all} should remain constant, as consistent perception under consistent stimuli is required. In other words, no discriminability should be found. Conversely, if the goal is to align with human perception, the graph shapes should mirror the discriminability observed in humans. 

However, several challenges arise with this approach in terms of reliability.
First, hyperparameters for the Gaussian filter should be chosen carefully. Excessive smoothing and noise reduction might lead to missing peaks in the original graph, potentially overlooking significant details.
Additionally, when analyzing peaks in the graph, the threshold for distance between embeddings was set arbitrarily, leading to a subjective inspection of peaks.
For instance, in the area under consideration in \autoref{fig:smooth_distance}, multiple peaks are evident and appear regular, yet the interpretation of each peak remains unclear.

Given that no studies have deeply investigated how similar these results are to human perception, our future work could involve detailed human studies. We believe that more thorough comparisons could be made to evaluate how closely the graph shapes resemble those humans perceive rather than focusing solely on peak analysis.

\subsection{General Benchmark for Channel Effectiveness} \label{sec:metric}

Our current evaluation of channel effectiveness primarily focuses on the accuracy and discriminability of embeddings across six different magnitude channels.
However, a comprehensive assessment of channel effectiveness should include additional metrics such as separability, popout, and grouping, which are crucial for understanding pre-attentive processing and just-noticeable difference (JND) in graphical perception.

To address these gaps, we propose developing a framework that measures model performance at a low level across these various metrics.
Our framework can be extended as a standardized benchmark that evaluates these fundamental aspects of graphical perception and provides a platform for comparing different visual encoders.
This benchmark would allow for a deeper understanding of how various models interpret and process graphical data, paving the way for advancements in chart comprehension technologies.

%% file: sections/06-Conclusion.tex
\section{Conclusion}

Our work introduces a novel framework for evaluating the graphical perception of image embedding models, focusing on the concept of channel effectiveness.
Our comprehensive experiments using the CLIP model have revealed significant disparities in how vision models and humans perceive and interpret visual channels. 
We observed that the accuracy and discriminability of these channels differ markedly between the CLIP model and human perception, suggesting the careful use of image embedding models on perception-related tasks.
Our findings highlight the potential for improving model reliability in tasks requiring human-like perception and precision, such as chart question answering and captioning.
As a future work, we suggest extending our benchmark to assess other low-level channel effectiveness, enhancing the robustness and reliability of visual encoders across diverse applications.
This approach not only promises reliable models in terms of graphical perception but also paves the way for future innovations in graphical data interpretation and machine learning in visualization.